\documentclass{article}

\usepackage{arxiv}

\usepackage[utf8]{inputenc} 
\usepackage[T1]{fontenc}    
\usepackage{hyperref}       
\usepackage{url}            
\usepackage{booktabs}       
\usepackage{amsfonts}       
\usepackage{nicefrac}       
\usepackage{microtype}      
\usepackage{cleveref}       
\usepackage{lipsum}         
\usepackage{graphicx}
\usepackage[numbers]{natbib}
\usepackage{doi}
\usepackage{listings}
\usepackage{xcolor}

\definecolor{codebg}{RGB}{245,245,245}

\title{Virtual Process Dossier: A Process-Aware Data Catalog}

\newif\ifuniqueAffiliation
\uniqueAffiliationfalse

\ifuniqueAffiliation 
\author{ \href{https://orcid.org/0009-0008-1828-5207}{\includegraphics[scale=0.06]{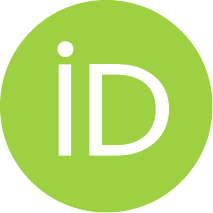}\hspace{1mm}Lukas Kubelka}\thanks{Corresponding author} \\
	Institute of Applied Informatics and Formal Description Methods (AIFB)\\
	Karlsruhe Institute of Technology\\
	Karlsruhe, Germany \\
	\texttt{lukas.kubelka@kit.edu} \\
}
\else
\usepackage{authblk}

\newbox{\orcid}\sbox{\orcid}{\includegraphics[scale=0.06]{orcid.pdf}} 
\author[1]{%
	\href{https://orcid.org/0009-0008-1828-5207}{\usebox{\orcid}\hspace{1mm}Lukas Kubelka}%
}
\author[2]{%
	\href{https://orcid.org/0000-0001-7866-8244}{\usebox{\orcid}\hspace{1mm}Alexander Bott}%
}
\author[3,6]{%
	\href{https://orcid.org/0009-0009-4323-9656}{\usebox{\orcid}\hspace{1mm}Frank D\"ohner}%
}
\author[6]{%
	\href{https://orcid.org/0009-0003-7820-3270}{\usebox{\orcid}\hspace{1mm}Saksham Kiroriwal}%
}
\author[4,7]{%
	\href{https://orcid.org/0009-0008-6947-127X}{\usebox{\orcid}\hspace{1mm}Georg Zeeb}%
}
\author[1,5]{%
	\href{https://orcid.org/0009-0003-5066-8412}{\usebox{\orcid}\hspace{1mm}Julia Butte}%
}
\author[6]{%
	Julius Pfrommer%
}
\author[6]{%
	\href{https://orcid.org/0000-0003-3556-7181}{\usebox{\orcid}\hspace{1mm}J\"urgen Beyerer}%
}
\author[1]{%
	\href{https://orcid.org/0000-0003-0576-7457}{\usebox{\orcid}\hspace{1mm}Tobias K\"afer}%
}
\affil[1]{Institute of Applied Informatics and Formal Description Methods, Karlsruhe Institute of Technology, Karlsruhe, Germany}
\affil[2]{Institute of Production Sciences, Karlsruhe Institute of Technology, Karlsruhe, Germany}
\affil[3]{Institute for Anthropomatics and Robotics, Karlsruhe Institute of Technology, Karlsruhe, Germany}
\affil[4]{Institute of Vehicle System Technology, Karlsruhe Institute of Technology, Karlsruhe, Germany}
\affil[5]{Institute of Information Security and Dependability, Karlsruhe Institute of Technology, Karlsruhe, Germany}
\affil[ ]{\small\texttt{\{firstname.lastname\}@kit.edu}\vspace{4pt}}
\affil[6]{Fraunhofer Institute of Optronics, System Technologies and Image Exploitation, Karlsruhe, Germany}
\affil[ ]{\small\texttt{\{firstname.lastname\}@iosb.fraunhofer.de}\vspace{4pt}}
\affil[7]{Fraunhofer Institute for Chemical Technology, Pfinztal, Germany}
\fi

\renewcommand{\shorttitle}{Virtual Process Dossier}

\hypersetup{
pdftitle={Virtual Process Dossier: A Process-Aware Data Catalog},
pdfsubject={cs.AI},
pdfauthor={Lukas Kubelka, Alexander Bott, Frank Doehner, Saksham Kiroriwal, Georg Zeeb, Julia Butte, Julius Pfrommer, Juerger Beyerer and Tobias Kaefer},
pdfkeywords={Data Catalog, Workflow Provenance, FAIR, Industry 4.0},
}

\begin{document}
\maketitle

\begin{abstract}
We propose the \textit{Virtual Process Dossier} (VPD), a Knowledge Graph-based data catalog that also captures workflow provenance.
We developed VPD for multi-stage manufacturing use-cases where downstream AI-based optimization tasks require to distinct between datasets generated during individual workflow steps.
VPD provides these datasets in a FAIR manner and makes both prospective and retrospective workflow provenance explicit.
Our contributions are:
(1) the VPD ontology that serves as the catalog's semantic core;
(2) the VPD provenance framework that integrates ontology instantiation into the production environment; and
(3) the VPD user interface that provides human-centered interaction with the VPD Knowledge Graph.
\textbf{VPD ontology:} \url{http://purl.org/vpd/vocab}\\
\textbf{VPD UI:} \url{https://github.com/kubeluk/VirtualProcessDossier}\\\textbf{License:} MIT
\end{abstract}

\section{Introduction}
Applying data-driven AI methods in manufacturing environments remains a significant challenge as data scientist face challenges to reliably trace where data originated and how it should be interpreted.
For engineering corresponding AI systems with the right data for the right analyses it is important to know, e.g. (i) what sensor data was generated during the manufacturing process of a product, (ii) where and how this data can be accessed, and (iii) what machines were involved in the production and how these were configured.
(iv) how did the overall production workflow look like, (v) during which of these steps was what sensor recording captured, and (vi) which machine was responsible for each step.
Such questions are hard to answer without management of metadata, which is on top tedious to reconstruct if it is not captured automatically.
As a result, manufacturing data is typically underutilized and remains overwhelming, rather than an easily reusable knowledge asset.
This paper presents the \textit{Virtual Process Dossier} to tackle the necessary information life-cycle management.

The issue of underutilized value of data is not limited to the manufacturing domain. Similar issues exist in larger projects or whole companies~\cite{jahnke2023catalog}.
We agree with \cite{groeger2021data,labadie2020fair,dinter2015metadata}, that this stems from the lack of a proper (meta)data management framework in place and leads to data being locked in silos.
To address this dilemma, a common proposal is to make data \underline{F}indable, \underline{A}ccessible, \underline{I}nteroperable, and \underline{R}eusable (FAIR)~\cite{wilkinson2016fair}.
The FAIR principles imply that unlocking the full value of data requires more than merely informing potential consumers of its existence or providing mechanisms for access.
To make effective use of data, consumers must be able to seamlessly combine it with other relevant data and, most importantly, understand its meaning.
To properly interpret the meaning, \textit{provenance} plays an important role~\cite{herschel2017survey}.

In previous works, data catalogs have emerged as a metadata management solution in academia and practice~\cite{ehrlinger2021catalog,schilling2020cdo}. Yet, current data catalog approaches seem to struggle when it comes to fostering FAIR data practices~\cite{labadie2020fair}.
While many approaches indeed provide features for describing the meaning of data, they merely focus on generic metadata like taxonomies or business glossaries~\cite{kropshofer2025catalog,labadie2020fair}.
This is not sufficient in a workflow-driven setting like ours, where detailed provenance about workflow executions is required.
Although ontologies have been identified as beneficial for domain modeling~\cite{failmayr2016ontology} and there exist well-established standards~\cite{w3c-ssn-sosa,w3c-prov-o,dcat-ap}, the majority of approaches do not leverage them~\cite{kropshofer2025catalog}.
A few exceptions are~\cite{carriero2025pko,verdejo2012p-plan,markovic2019ep-plan,dibowski2020using}.
However, the proposed solutions usually either focus solely on digital‑data generation processes or fail to suit multi‑stage manufacturing environments, neglecting the linkage of datasets to individual, stage‑specific provenance information.

\begin{figure}[t]
\centering
\includegraphics[width=1.0\textwidth]{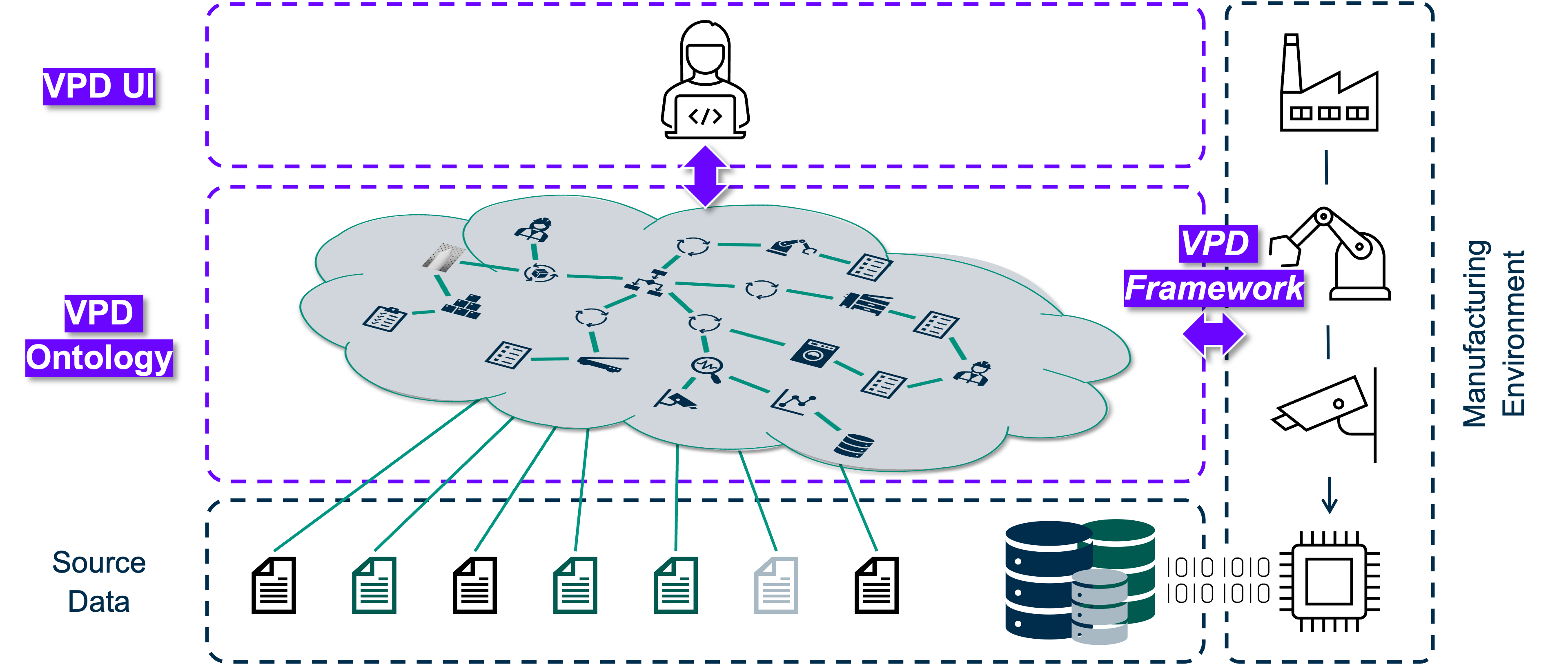}
\caption{The VPD is our approach to FAIR data management in multi-stage workflow domains such as manufacturing.}
\label{fig:kg-sem-search}
\end{figure}

We propose the \textit{Virtual Process Dossier} (VPD), a knowledge graph-based, and process-aware data catalog for multi-stage workflow environments.
These environment produces physical entities while generating vast amounts of sensor observations throughout workflow executions which are stored in various, external data sources.
The VPD constitutes a \textit{knowledge layer} that sits above the raw \textit{source data} layer of sensor observations -- i.e.\ it semantically augments primary data, which is stored elsewhere, in order to make it FAIR.
The VPD is comprised of the components highlighted in Fig. \ref{fig:kg-sem-search} which represent our main contributions:
\begin{enumerate}
    \item The \textit{VPD ontology}\footnote{\url{http://purl.org/vpd/vocab}} represents the knowledge layer in our approach. It semantically describes the workflow provenance of physical, multi-stage production processes and links it to digital data generation processes happening in between by aligning the DCAT~\cite{w3c-dcat}, WiLD~\cite{kaefer2018wild}, PROV~\cite{w3c-prov-o} and SSN/SOSA~\cite{w3c-ssn-sosa} ontologies.
    \item The \textit{VPD framework} provides guiding steps on how the VPD knowledge graph is instantiated based on the VPD ontology and how it can be integrated into a manufacturing environment for capturing provenance during workflow executions and linking it to generated sensor datasets automatically.
    \item The \textit{VPD UI}\footnote{\url{https://github.com/kubeluk/VirtualProcessDossier}} is a Web application that sits logically above the VPD knowledge graph. It provides a guided way for non-expert users to query, browse and update the VPD by visualizing datasets and their provenance.
\end{enumerate}

The rest of the paper is structured as follows:
We first introduce preliminaries in section \ref{sec:prelim}, followed by the running example in section \ref{sec:example}.
In section \ref{sec:vpd-o}, we present the VPD ontology.
Afterwards, we explain the accompanying framework for constructing the VPD knowledge graph in section \ref{sec:framework}.
Section \ref{sec:system} provides a brief overview over the VPD UI's tech stack.
Finally, we state related work in section \ref{sec:related} and conclude this paper in section \ref{sec:conclusion}.

\section{Preliminaries} \label{sec:prelim}

In this section we provide a brief overview over the FAIR principles and introduce relevant provenance terminology that we use throughout the paper.

\subsection{FAIR Principles} \label{prelim:fair}

The FAIR data principles~\cite{wilkinson2016fair} have been proposed by a diverse set of stakeholders for providing guidelines to (meta)data management and stewardship. They are composed of the four core principles \textit{Findable}, \textit{Accessible}, \textit{Interoperable} and \textit{Reusable}, which themselves were respectively refined into sub-principles. In summary, they promote (meta)data to be globally identifiable and searchable (F), by which it is retrievable using a standardized communication protocol (A). It should use a broadly applicable knowledge representation language (I) -- e.g. RDF~\cite{rdf-concepts}. Finally, detailed provenance information and usage conditions should be present (R).

\subsection{Provenance} \label{prelim:prov}

Provenance has been extensively studied across domains and is generally defined as any information describing the production process of an end product -- digital or physical~\cite{herschel2017survey}. As such, it comprises ``meta-data about entities, data, processes, activities, and persons involved in the production process''~\cite{groth2013prov}.
In general, four different types of provenance can be distinguished that decrease in scope of what kind of provenance they collect~\cite{herschel2016provenance}: (i) provenance meta-data; (ii) information system provenance; (iii) workflow provenance; and (iv) data provenance. Workflow provenance focuses on production processes that can be represented as workflows -- e.g. through a directed graph, where nodes represent units of work and edges represent data or control flow. It can be further divided into (iii.a) prospective provenance; (iii.b) retrospective provenance; and (iii.c) evolution provenance. Prospective provenance captures an abstracted overview of a workflow, which is independent of its execution~\cite{alper2013enhancing}. Retrospective provenance captures information related to a workflow's execution. Evolution provenance relates to capturing changes between versions of a workflow.
In this paper, we view manufacturing processes as workflows that model the control flow of production steps.

\subsection{DCAT, PROV, SSN/SOSA and WiLD ontology} \label{prelim:ontologies}

\textit{DCAT}~\cite{dcat-ap} provides a standardized vocabulary for describing datasets, their distributions, and access mechanisms and supports interoperable metadata exchange.
\textit{PROV}~\cite{w3c-prov-o} offers standardized terminology for modeling retrospective provenance through entities, activities, and agents, enabling the representation of how artifacts were produced, transformed, or used.
\textit{SSN/SOSA}~\cite{w3c-ssn-sosa} supplies a standardized framework for describing sensors and their observations, including the relations between sensing devices, observed properties, and generated results.
\textit{WiLD}~\cite{kaefer2018wild} introduces vocabulary for stating retrospective as well as prospective workflow provenance. It comes with an operational semantics which makes it executable, e.g. by ASM4LD~\cite{kaefer2018asm4ld}.
Together, these ontologies provide interoperable building blocks for expressing dataset metadata, provenance, and sensor observations in a standardized or published way.

\section{Running Example} \label{sec:example}

\begin{figure}[ht]
\centering
\includegraphics[width=0.85\textwidth]{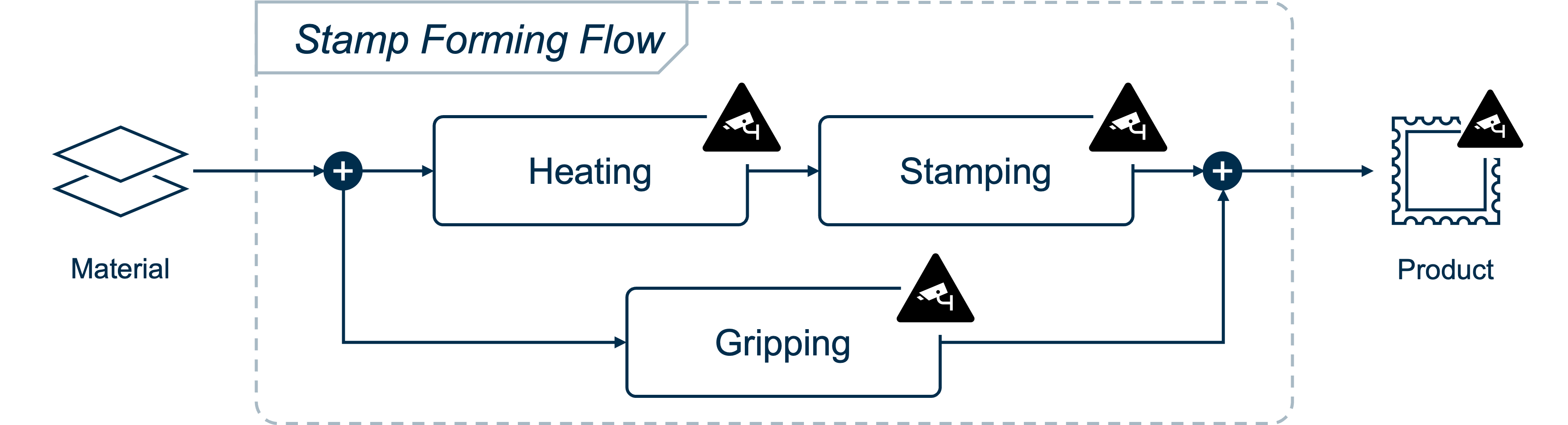}
\caption{The multi-step manufacturing workflow presenting our running example}
\label{fig:stamp-forming-flow}
\end{figure}

We use a simplified production process throughout this paper as a running example. It essentially describes a two-step manufacturing workflow for producing fiber-reinforced plastic parts by, starting from an initial sheet of fiber material, (i) heating it in an oven and; (ii) forming it with a press. Throughout the process the sheet is held in place by a gripper frame in which it travels from step to step. There are several sensors involved in this process that observe various feature properties like surface temperatures or gripper angles during runtime.
This process is presented in Fig.\ \ref{fig:stamp-forming-flow}. It resembles in this case a closer description of the manufacturing environment mentioned in Fig.\ \ref{fig:kg-sem-search}.

\section{VPD Ontology} \label{sec:vpd-o}

This section introduces the VPD ontology, which represents the data catalog's semantic core. It provides the basis for making sensor data in our use-case FAIR by providing the structure of how sensor data is linked to prospective and retrospective provenance and thus how it is intended to be interpreted. Section \ref{subsec:competency} explains our approach to ontology engineering. Section \ref{subsec:ontology} elaborates on the ontology's structure.

\subsection{Ontology Engineering} \label{subsec:competency}

We conducted multiple workshops with VPD stakeholders from our manufacturing use-case -- process engineers, mechanical engineers and data scientists. The goal was to extract the individual information needs of each group. This involved building a shared understanding of the manufacturing environment, e.g. which products would be produced, what the possible production workflows would look like, what sensors and machines could be deployed, and how down-stream AI tasks may look like. From several group discussions and individual refinements, \textit{competency questions} such as the following emerged which describe the needed forms of knowledge about workflow provenance for said down-stream tasks:
\begin{enumerate}
    \item Which data $D_1, \dots, D_n$ was produced during production of product $P$?
    \item How did the production workflow $W$ look like for $P$?
    \item During which step in $W$ was data $D_i$ recorded?
    \item What was the observation target $T$ and observed aspect $A$ for dataset $D_j$?
    \item What happened during, before or after the observations of $D_i$?
    \item Which data $D_1, \dots, D_j$ are observations of machine $M_k$ or product $P_l$?
    \item Which data $D_1, \dots, D_j$ come from runs where first machine $M_1$ with value $V_1$ for parameter $C_1$ performed an action and afterwards machine $M_2$ with value $V_2$ for parameter $C_2$?
\end{enumerate}

After the collection of competency questions, we researched published, well-known and standardized ontologies that are applicable for our use-case. Our aim was to re-use existing vocabularies as frequently as possible, since this is considered to be best practice.
As a result, the VPD ontology largely comprises terms that come from published or standardized ontologies -- i.e. DCAT~\cite{dcat-ap}, PROV~\cite{w3c-prov-o}, SOSA/SSN~\cite{w3c-ssn-sosa}, WiLD~\cite{kaefer2018wild}.
It further only introduces a few new classes for the sake of being used as alignment elements between the listed ontologies.
We accessed the ontology's fairness using the FOOPS! tool\footnote{\url{https://foops.linkeddata.es/FAIR\_validator.html}}.

\subsection{Ontology Overview} \label{subsec:ontology}

We structure our explanation about VPD-O in three parts that respectively present a certain view on the ontology: (i) Sensors and datasets, (ii) machines and parameters, and (iii) products and materials.

\subsubsection{Sensors and Datasets.}

\begin{figure}[]
        \centering
        \includegraphics[width=0.7\linewidth]{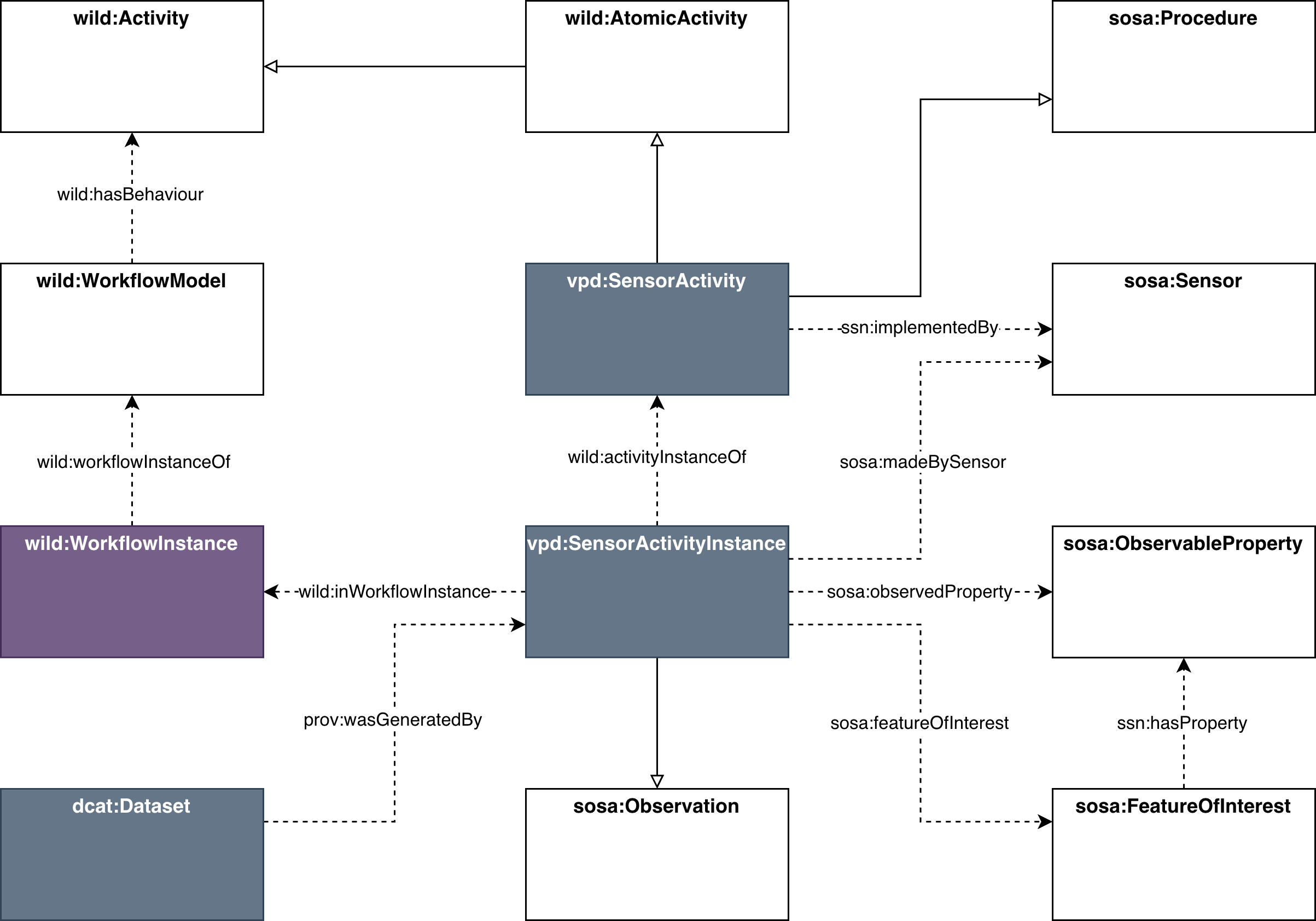} 
        \caption{VPD-O view on sensor activities.}
        \label{fig:vpd-o-sensor}
\end{figure}
    
\begin{figure}[]
        \centering
        \includegraphics[width=0.7\linewidth]{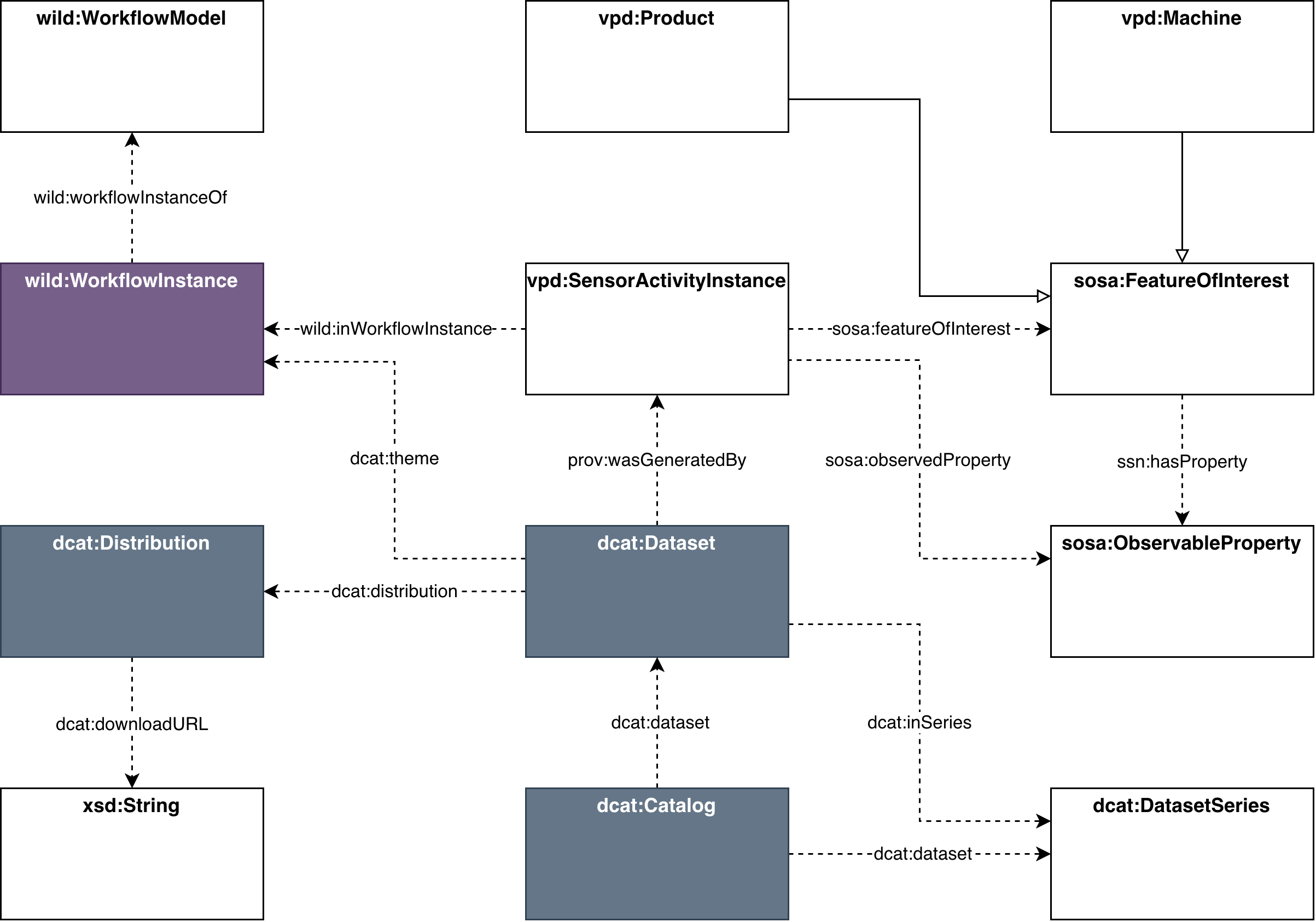}
        \caption{VPD-O view on datasets.}
        \label{fig:vpd-o-catalog}
\end{figure}

As the competency questions suggest, we are foremost interested in questions about sensor data produced during multi-stage production workflows and their related provenance. The PROV~\cite{w3c-prov-o} ontology is limited in this regard and does not enable to model parallel activities or the topological relation between activities explicitly, since it was developed for capturing \textit{retrospective} provenance. Identifying activities that happened simultaneously is possible by comparing their start (\texttt{prov:startedAtTime}) and end times (\texttt{prov:endedAtTime}), but statements about the order and general flow of activities cannot be made as this requires terminology for describing \textit{prospective} provenance. Such terms are provided by the WiLD~\cite{kaefer2018wild} ontology -- e.g. due to its distinction between \texttt{wild:AtomicActivity}, \texttt{wild:SequentialActivity} and \texttt{wild:ParallelActivity}.
As hinted in Fig. \ref{fig:vpd-o-sensor}, a workflow in WiLD is represented as a type of \texttt{wild:WorkflowModel} which represents the static, abstracted process. A workflow model has a root behaviour (\texttt{wild:hasBehaviour}) that builds the root activity of the tree-based workflow structure. At some point in this hierarchy there will be instances of type \texttt{wild:AtomicActivity} that are the individual actions to be performed by machines and sensors.
The running example's workflow model could be described as such\footnote{Due to space constraints we omitted the further triples for the gripping and forming activities as well as all atomic activities such as further sensor activities.}:

\begin{lstlisting}
@prefix wild: <http://purl.org/wild/vocab#> .
@base <http://example.org/> .

<wf/1#it> a wild:WorkflowModel ;
    wild:hasBehaviour <wf/1/root#it> .

<wf/1/root#it> a wild:ParallelActivity ;
    wild:hasChildActivities (
        <wf/1/root/heatingAndForming#it>
        <wf/1/root/gripping#it>
    ) .

<wf/1/root/heatingAndForming#it> a wild:SequentialActivity ;
    wild:hasChildActivities (
        <wf/1/root/heatingAndForming/heating#it>
        <wf/1/root/heatingAndForming/forming#it>
    ) .

<wf/1/root/heatingAndForming/heating#it> a wild:ParallelActivity ;
    wild:hasChildActivities (
        <wf/1/root/heatingAndForming/heating/do#it>
        <wf/1/root/heatingAndForming/heating/sense/upperSurfaceTemperature#it>
        <wf/1/root/heatingAndForming/heating/sense/lowerSurfaceTemperature#it>
    ) .
\end{lstlisting}

In VPD-O, as shown in Fig. \ref{fig:vpd-o-sensor}, we introduce \texttt{vpd:SensorActivity} as a subclass of \texttt{wild:AtomicActivity} and \texttt{sosa:Procedure}. When a workflow model is executed, basically two things happen: (1) a \texttt{wild:WorkflowInstance} is created, representing the workflow execution, and (2) all activities contained in the workflow model's behavior get instantiated by a respective \texttt{wild:ActivityInstance}.
The corresponding counterpart for our sensor activities is modeled by the class \texttt{vpd:SensorActivityInstance} -- i.e. it is a subclass of \texttt{wild:ActivityInstance}. As it is used to describe retrospective sensor activities it is further subclass of \texttt{sosa:Observation}\footnote{The introduction of a dedicated subclass is necessary, since \texttt{wild:ActivityInstance} has an attached state and represents completed activites (i.e. \texttt{prov:Activity}) only when their state is \texttt{wild:done}.}.
From this connection it is possible to add further information about the sensor activity instance like \texttt{sosa:FeatureOfInterest}, its \texttt{sosa:ObservableProperty} and the \texttt{sosa:Sensor} that was responsible.
For the observation of the product's upper surface temperature during the heating step from our running example, this could look like the following:

\begin{lstlisting}
<wfI/1#it> a wild:WorkflowInstance ;
    wild:workflowInstanceOf <wf/1#it> .

<wfI/1/root/heatingAndForming/heating/sense/1#it> a wild:SensorActivityInstance, sosa:Observation ;
    wild:inWorkflowInstance <wfI/1#it> ;
    wild:activityInstanceOf <wf/1/root/heatingAndForming/heating/sense/1#it> ;
    sosa:madeBySensor <sensors/pyrometer/1#it> ;
    sosa:featureOfInterest <product/1#it> ;
    sosa:observedProperty <product/1#upperSurface> .
\end{lstlisting}

The SSN/SOSA ontology provides a PROV alignment module which states that \texttt{sosa:Observation} is a subclass of \texttt{prov:Activity}. This also applies to our sensor activity instance through the transitivity of the subclass relationship. From that, the possibility for linking a \texttt{dcat:Dataset} to a sensor observation emerges.
The DCAT vocabulary states the \texttt{prov:wasGeneratedBy} property which can be used to describe the \texttt{prov:Activity} that lead to the generation of a \texttt{dcat:Dataset}.

Based on this simple alignment between WiLD, PROV, SSN/SOSA and DCAT, the VPD ontology enables to describe the generation process of sensor data by stating in which explicit step in a workflow execution it was generated, which sensor was responsible for it and which entity and according property was observed.
VPD-O immediately catalogs all datasets as part of a \texttt{dcat:Catalog} that represents the VPD data catalog (cf. Fig. \ref{fig:vpd-o-catalog}). Datasets are further directly linked to the workflow execution (\texttt{wild:WorkflowInstance}) via the \texttt{dcat:theme} property\footnote{The reason behind this is simply w.r.t. simplifying SPARQL queries looking for all datasets related to a production run.}. The \texttt{dcat:Distribution} is used to state where and how a dataset may be accessed.
Applied to our example, this could look like so:

\begin{lstlisting}
<vpd/cat#it> a dcat:Catalog ;
    dcat:dataset <vpd/ds/1#it> .

<vpd/ds/1#it> a dcat:Dataset ;
    dcat:theme <wfI/1#it> ;
    dcat:distribution [
        a dcat:Distribution ;
        dcat:downloadURL <http://store.example.org/path/to/file.csv>
    ] ;
    prov:wasGeneratedBy <wfI/1/root/heatingAndForming/heating/sense/1#it> .
\end{lstlisting}

\subsubsection{Machines and Parameters.}

\begin{figure}[]
    \centering
    \includegraphics[width=0.7\linewidth]{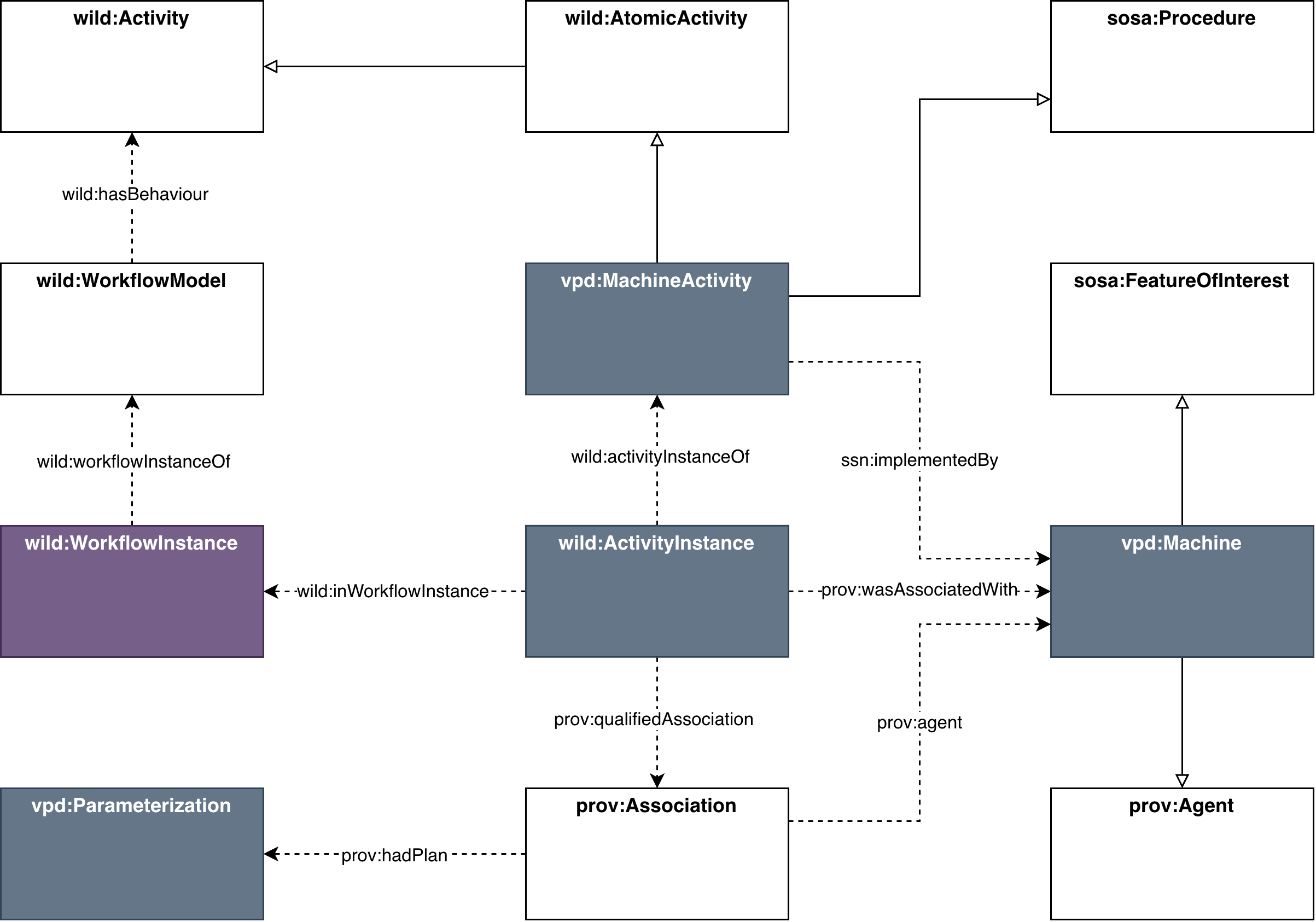} 
    \caption{VPD-O view on machine activities.}
    \label{fig:vpd-o-machine}
\end{figure}

Next, from the competency questions we can take that we are further interested in (i) observations regarding machines, and (ii) how they were parameterized.
Thus, Fig. \ref{fig:vpd-o-machine} shows similar ideas from our elaborations about sensors in the workflow. We again have a dedicated activity class -- \texttt{vpd:MachineActivity} -- that is used for alignment with WiLD and SSN/SOSA as a sublass of \texttt{wild:AtomicActivity} and \texttt{sosa:Procedure}. For capturing the parameterizations for machines we introduce the class \texttt{vpd:Machine} as a subclass of \texttt{prov:Agent} and the class \texttt{vpd:Parameterization}. For associating parameterizations with machines, we apply the qualification pattern~\cite{dodds2022patterns} used in PROV. When a machine performed a task, we associate it with the corresponding \texttt{wild:ActivityInstance} via the \texttt{prov:wasAssociatedWith} property. We then qualify this association and attach the parameterization which we model as a \texttt{prov:Plan}. For representing individual parameter values, we utilize the QUDT ontology~\cite{qudt}. The provenance for the heating activity from our running example could be described as such:

\begin{lstlisting}
@prefix qudt: <http://qudt.org/schema/qudt/> .
@prefix unit: <http://qudt.org/vocab/unit/> .

<wfI/1/root/heatingAndForming/heating/do#it> a wild:ActivityInstance ;
    wild:inWorkflowInstance <wfI/1#it> ;
    wild:activityInstanceOf <wf/1/root/heatingAndForming/heating/do#it> ;
    prov:wasAssociatedWith <machines/oven/1#it> ;
    prov:qualifiedAssociation [
        a prov:Association ;
        prov:agent <machines/oven/1#it> ;
        prov:hadPlan [
            a prov:Plan ;
            vpd:heatingTemperature [
                a qudt:QuantityValue ;
                qudt:value 200.0 ;
                unit:DEG_C
            ]
        ]
    ] .
\end{lstlisting}

Sensors also observe the activities of machines in our use-case. Up to this point, with VPD-O it is possible to query which machine activities happened before, during or after a certain sensor observation via \texttt{wild:inWorkflowInstance} and \texttt{wild:activityInstanceOf} which link to the static, abstracted workflow description.
Though, it is possible that during one machine activity there could be multiple sensors involved that observe its execution. For such cases, it would not be possible to separate the generated dataset from each other.
In order to make this distinction clear, we further model \texttt{vpd:Machine} as a subclass of \texttt{sosa:FeatureOfInterest}. As shown in Fig. \ref{fig:vpd-o-parameter}, this enables to attach multiple \texttt{sosa:ObservableProperty} to it and describe observation-relevant aspects. In our running example this could be e.g. the angles and forces of grippers inside the gripper frame during the forming step.

Figure \ref{fig:vpd-o-parameter} also presents that we use SHACL~\cite{shacl-spec} shapes in the form of the \texttt{vpd:ParameterDescription} class as descriptions for how certain machine parameterizations need to be represented\footnote{The reason for modeling with shapes is for utilizing them for the automated generation of HTML forms. More information on this will be provided in section \ref{sec:system}}.
Within these shapes, parameters are referenced via an individual \texttt{sh:PropertyShape} and further described by an individual \texttt{sh:NodeShape} which simultaneously represents a property that may be observed by sensors.
For the press in our example this may look like this:

\begin{lstlisting}
@prefix sh: <https://www.w3.org/ns/shacl> .

vpd:drapingPressParameterDescription a sh:NodeShape, vpd:ParameterDescription ;
    sh:targetClass vpd:DrapingPressParametrization ;
    sh:property [
        sh:path fofa:pressForce ;
        sh:name "Force" ;
        sh:node fofa:pressForceParameter ;
        sh:minCount 1 ;
        sh:maxCount 1
    ] .

vpd:pressForceParameter a sh:NodeShape, sosa:ObservableProperty ;
    sh:targetClass vpd:PressForce ;
    sh:property [
        sh:path qudt:hasUnit ;
        sh:name "Unit" ;
        sh:in ( unit:N ) ;
        sh:defaultValue unit:N ;
        sh:nodeKind sh:IRI ;
        sh:maxCount 1 ;
        sh:minCount 1
    ] ;
    sh:property [
        sh:path qudt:value ;
        sh:name "Value" ;
        sh:minCount 1 ;
        sh:maxCount 1 ;
        sh:minInclusive 60 ;
        sh:maxInclusive 900 ;
        sh:defaultValue 100 ;
        sh:datatype xsd:integer
    ] .
\end{lstlisting}

\subsubsection{Products and Materials.}

Finally, we provide the ability to describe that a certain product was the output of a respective workflow execution. In Fig. \ref{fig:vpd-o-product} there is the class \texttt{vpd:Product} that is connected a \texttt{vpd:Material} which it was derived from. Again, for alignment with PROV, both of these classes are subclasses of \texttt{prov:Entity}. Like we did before, we apply the qualification pattern to this relation and link workflow executions (\texttt{wild:WorkflowInstance}) via \texttt{prov:Derivation} as a corresponding \texttt{prov:Generation}.

\section{VPD Framework} \label{sec:framework}

The VPD-O comes with a framework that provides a straight-forward way for describing various workflow-based manufacturing environments and capturing the provenance of workflow runs -- i.e. a guide to instantiate and populate the VPD knowledge graph according to the ontology. It comprises three phases that roughly divide the required actions according to whether they happen before (\ref{subsec:before}), during (\ref{subsec:during}) or after (\ref{subsec:after}) the manufacturing of products.

\subsection{Setup and Initial Handshake} \label{subsec:before}

Initially, before any production run may happen, apply the following steps for setting up the VPD knowledge graph and priming it with prospective workflow provenance. This preparation work is crucially important as the later phases are reliant on a tight coupling and coordination between the manufacturing system and the knowledge graph in order to capture retrospective workflow provenance:

\begin{enumerate}
    \item Instantiate \texttt{dcat:Catalog} by minting the fresh IRI~\cite{rfc3987iri} $V^P_D$.
    \item Instantiate \texttt{vpd:Machine} and \texttt{sosa:Sensor} for every system in the environment by minting the fresh IRIs $M_i$ and $S_j$.
    \item Describe the systems $M_i$ and $S_j$ using the SSN/SOSA~\cite{w3c-ssn-sosa} vocabulary as needed.
    \item Instantiate \texttt{sosa:ObservableProperty} with IRI $O_k$ for every aspect that will be observed throughout production runs. Link each entity with its respective property which will be observed via \texttt{ssn:hasProperty}.
    \item Describe the manufacturing workflow using the WiLD~\cite{kaefer2018wild} vocabulary by instantiating \texttt{wild:WorkflowModel} with IRI $W$ and link the IRIs $A_l$ for instances of type \texttt{wild:AtomicActivity} to the implementing system via \texttt{ssn:implementedBy}.
    \item For each system, distribute the IRI tuple $(A_l,M_i)$ to the respective controller $C_m$ that is responsible for its activation. If $A_l$ is a sensor activity, also attach $(V^P_D,S_j,O_k)$. In simple terms: Every actor in the physical manufacturing environment must possess a name by whom it is globally identifiable and must be aware of the activity's name which it will be executing. Further, sensors must be aware of the data catalog's name and which property of what entity it will be observing.
\end{enumerate}

\subsection{Final Handshake and Execution} \label{subsec:during}
    
When a new production run is scheduled and executed, apply the following steps for capturing retrospective provenance of machine and sensor activities (synchronously and in a distributed manner or asynchronously in a local manner, or in a combination): (i) the production environment immediately inserts retrospective provenance information into the VPD knowledge graph using an event-driven approach as actions are performed and completed, or (ii) the execution context is collected within the manufacturing management system and inserted in bulk from a central controller:

\begin{enumerate}
    \item Instantiate the workflow model $W$ by constructing a new workflow instance IRI $W^I$ for itself and activity instance IRIs $A^I_{c}$ for all attached activities in the workflow tree, except for the atomic activities $A^I_a$ which are the machine and sensor activities. Those will be instantiated by the manufacturing system on-demand and augmented with the active runtime provenance information which we will explain next.
    \item Publish the generated workflow instance IRI $W^I$ to each controller $C_m$ that manages the execution of $A_l$.
\end{enumerate}

At this stage, all relevant IRIs have been distributed to all $C_m$ which marks the completion of the synchronization process (two-way ``\textit{handshake}'') between the manufacturing management system and the VPD knowledge graph that is required for capturing provenance information.
From this point on, the manufacturing system executes in a distributed, open-loop fashion and updates the VPD knowledge graph by capturing provenance information as events occur.
This is simply done through directly sending SPARQL/\texttt{UPDATE}~\cite{sparql-update} queries from the manufacturing environment to the triple store where the VPD knowledge graph is hosted. We will explain this step for sensor and machine activity executions next.

\subsubsection{Sensors executions.} For any execution $A^I_a$ that instantiates a sensor activity \texttt{vpd:SensorActivity}, the responsible controller $C_m$ collects the sensor reading dataset and mints the IRI $D$ for it. It then adds it to the data catalog and links it to $W^I$. The following SPARQL/Update template does this:

\begin{lstlisting}[mathescape]
INSERT DATA
{ 
  <$A^I_a$> a vpd:SensorActivityInstance, wild:ActivityInstance, sosa:Observation ;
    wild:activityInstanceOf <$A_l$> ;
    wild:inWorkflowInstance <$W^I$> ;
    sosa:madeBySensor <$S_j$> ;
    sosa:observedProperty <$O_k$> ;
    sosa:featureOfInterest <$M_i$> .
  <$D$> a dcat:Dataset ;
    prov:wasGeneratedBy <$A^I_a$> .
  <$V^P_D$> dcat:dataset <$D$> .
}
\end{lstlisting}

$C_m$ also sends the dataset and the IRIs $D$ and $A^I_a$ to the a component $C^{'}_m$ that is responsible for storing the dataset persistently. This component then stores the dataset and acquires the identifier $L$. Finally, $C^{'}_m$ mints the IRI $G$ for a \texttt{dcat:Distribution} and sends the following SPARQL/Update query for linking $D$ with $G$ that uses $L$ as object for \texttt{dcat:downloadURL}:

\begin{lstlisting}[mathescape]
INSERT DATA
{ 
  <$G$> a dcat:Distribution ;
    dcat:downloadURL <$L$> .
  <$D$> dcat:distribution <$G$> .
}
\end{lstlisting}

\subsubsection{Machine executions.} For machine activity instances the steps taken for capturing the provenance information are analogous to the ones we explained for sensor activity instances and are thus omitted for brevity.

\subsection{Retrospective Augmentation} \label{subsec:after}

Other retrospective provenance information about the newly generated IRIs of production run $W^I$ may also be added to the knowledge graph. Reasons for doing this include: (i) The information cannot be automatically captured -- e.g. any form of relevant human observations that occurred throughout the manufacturing process, (ii) the procedure for doing so has not been implemented yet, or (iii) the procedure will not be implemented due to complexity or other reasons.

\section{VPD System} \label{sec:system}

Accompanying to the VPD-O and its instantiation framework we propose the VPD data catalog user interface (UI). The VPD UI adds an application layer on top of the semantic knowledge layer and is directly fed from the VPD knowledge graph that is hosted by a triple store. As such, it is intended for users that are unfamiliar with the VPD-O structure and constructing SPARQL queries for accessing graph itself. The VPD UI provides the following features: (i) register production runs, (ii) perform retrospective augmentation, (iii) browse and search for datasets, and (iv) visualize the provenance of production runs. We publish the UI's source code as an open-source software artifact.
As mentioned before, we leverage SHACL shapes for representing machine parameterizations. We use them to generating HTML forms which can be used to view, modify, generate and validate RDF triples from user inputs.
We utilize the shacl-form\footnote{\url{https://github.com/ULB-Darmstadt/shacl-form}} library for this means. It powers the UI view to browse and add machine parameterizations for workflow executions.

\section{Related Work} \label{sec:related}

The PROV ontology~\cite{w3c-prov-o} provides vocabulary that is applicable for capturing retrospective provenance. However, on its own it is not possible to describe the structural dependencies between occurred activities and thus to capture the static context of a workflow -- i.e. prospective provenance.

With D-PROV~\cite{missier2013dprov} there exists a proposal for extending PROV with means for describing prospective provenance. As opposed to our approach for capturing provenance information, it rather focuses on the production of digital data products instead of physical entities. As a result, workflow descriptions in D-PROV represent data flows which is why it is not applicable in our manufacturing use-case.

The PCPAC ontology presented in~\cite{dibowski2020using} is used to introduce a data catalog approach for establishing FAIR data management in data lake architectures. Although the approach captures provenance information, it merely captures information system provenance which states the introduction of datasets to the data lake or their modification. As such, DCPAC is not applicable for our manufacturing scenario.

The Procedural Knowledge Ontolgy (PKO)~\cite{carriero2025pko} presents the closest fit to our use case. It focuses on workflows that model the control flow of industrial companies and like WiLD~\cite{kaefer2018wild} provides means for distinguishing between abstracted activities and their actual executions -- i.e. prospective vs. retrospective provenance.
However, PKO takes a flow-based view on processes. The properties \texttt{pko:nextStep} and \texttt{pko:nextAlternativeStep} indeed provide means for describing the order of steps. Nevertheless, the latter cannot be used to denote parallelism. It can be thought of as a logical xor relationship\footnote{\url{https://w3id.org/pko\#nextAlternativeStep}}. Using \texttt{pko:nextStep} to describe parallelism would be a hack, as the documentation implies a 1:1 cardinality between the previous and the next step.
The WiLD ontology takes a tree-based view on processes, similar to OWL-S~\cite{owl-s}. In WiLD, child activities of \texttt{wild:Activity} are modeled as RDF lists, which gives them an order and an explicit closure. Sub-classes of \texttt{wild:Activity} determine how this list must be interpreted, e.g. \texttt{wild:SequentialActivity}, where activities are running in sequence (i.e. closure and order matters), or \texttt{wild:ParallelActivity}, where activities are running in parallel (i.e. only closure matters), or \texttt{wild:ConditionalActivity}, where there is a logical xor relationship (i.e. only closure matters and conditions need to be considered).
For a comparison of processes as flows and trees and their convertibility see the work of Vanhatalo et al~\cite{vanhatalo2009process}.

PKO internally uses the P-Plan ontology~\cite{verdejo2012p-plan} for representing workflow steps, that was extended by the EP-Plan~\cite{markovic2019ep-plan} ontology. EP-Plan was designed for modeling scientific workflows. Such workflows describe data dependencies~\cite{herschel2017survey} and hence are not applicable in our use-case.

\section{Conclusion} \label{sec:conclusion}

With the introduction of the \textit{Virtual Process Dossier} (VPD) we presented an approach to foster FAIR data management practices in multi-stage manufacturing environments by leveraging our \textit{VPD-O}ntology and Semantic Web technologies to create process-aware data catalogs.
\textit{VPD-O} reuses well-know, standardized and published ontologies and enables to describe the provenance of sensor data by leveraging the physical production workflow as a first-class citizen. It comes with a straight-forward framework for instantiating and populating the VPD knowledge graph, and scaling it to manufacturing workflows of arbitrary complexity.
For browsing the VPD catalog and visualizing captured provenance information, we introduced the VPD UI which we publish as an open-source software artifact.
VPD could have the potential to play a crucial role in various AI-driven manufacturing use-cases as it presents an extensible, scalable, interoperable and standards-based solution to break down data silos in this domain.

We built VPD for a research endeavour to improve immature manufacturing processes, where a lot of data around a manufacturing process is measured and simulated, and needs to be used in downstream analyses by data scientists.
This endeavour involves 8 research labs in Karlsruhe, Germany, spanning three faculties of the Karlsruhe Institute of Technology (KIT), next to two Fraunhofer institutes (IOSB and ICT).
As VPD solves a general problem, and we strived for a general solution, we believe it is useful for the larger community.

\bibliographystyle{splncs04}
\bibliography{references}

\begin{thebibliography}{10}
\providecommand{\url}[1]{\texttt{#1}}
\providecommand{\urlprefix}{URL }
\providecommand{\doi}[1]{https://doi.org/#1}

\bibitem{w3c-dcat}
Albertoni, R., Browning, D., Cox, S.J.D., et~al.: Data catalog vocabulary (dcat). W3C Recommendation REC-vocab-dcat-3, W3C (Aug 2024), \url{https://www.w3.org/TR/vocab-dcat-3/}

\bibitem{alper2013enhancing}
Alper, P., Belhajjame, K., Goble, C.A., et~al.: Enhancing and abstracting scientific workflow provenance for data publishing. In: Proceedings of the Joint EDBT/ICDT 2013 Workshops. p. 313–318. EDBT '13, Association for Computing Machinery, New York, NY, USA (2013). \doi{10.1145/2457317.2457370}

\bibitem{carriero2025pko}
Carriero, V.A., Scrocca, M., Baroni, I., et~al.: Procedural knowledge ontology (pko). In: The Semantic Web: 22nd European Semantic Web Conference, ESWC 2025, Portoroz, Slovenia, June 1–5, 2025, Proceedings, Part II. p. 334–350. Springer-Verlag, Berlin, Heidelberg (2025). \doi{10.1007/978-3-031-94578-6_19}

\bibitem{rdf-concepts}
Cyganiak, R., Wood, D., Lanthaler, M.: Rdf 1.1 concepts and abstract syntax. W3c recommendation, W3C (Feb 2014), \url{https://www.w3.org/TR/rdf11-concepts/}

\bibitem{dibowski2020using}
Dibowski, H., Schmid, S., Svetashova, Y., et~al.: Using semantic technologies to manage a data lake: Data catalog, provenance and access control. In: Proceedings of the 13th International Workshop on Scalable Semantic Web Knowledge Base Systems co-located with 19th International Semantic Web Conference (ISWC 2020). CEUR Workshop Proceedings, vol.~2757, pp. 65--80 (Nov 2020)

\bibitem{dinter2015metadata}
Dinter, B., Gluchowski, P., Schieder, C.: A stakeholder lens on metadata management in business intelligence and big data -- results of an empirical investigation. In: AMCIS 2015 Proceedings (2015)

\bibitem{dodds2022patterns}
Dodds, L., Davis, I.: Qualified relation. In: Linked Data Patterns: A pattern catalogue for modelling, publishing, and consuming Linked Data. Online (2022), \url{https://patterns.dataincubator.org/}

\bibitem{rfc3987iri}
Dürst, M.J., Suignard, M.: {Internationalized Resource Identifiers (IRIs)}. {RFC}~3987, IETF (Jan 2005), \url{https://www.rfc-editor.org/info/rfc3987}

\bibitem{ehrlinger2021catalog}
Ehrlinger, L., Schrott, J., Melichar, M., et~al.: Data catalogs: A systematic literature review and guidelines to implementation. In: Database and Expert Systems Applications - DEXA 2021 Workshops. pp. 148--158. Springer International Publishing (2021)

\bibitem{failmayr2016ontology}
Feilmayr, C., W\"{o}ß, W.: An analysis of ontologies and their success factors for application to business. Data \& Knowledge Engineering  \textbf{101},  1--23 (2016), \url{https://doi.org/10.1016/j.datak.2015.11.003}

\bibitem{verdejo2012p-plan}
Garijo~Verdejo, D., Gil, Y.: Augmenting prov with plans in p-plan: Scientific processes as linked data. In: Proceedings of the 2nd International Workshop on Linked Science. vol.~951. CEUR Workshop Proceedings (2012), \url{http://ceur-ws.org/Vol-951/}

\bibitem{sparql-update}
Gearon, P., Passant, A., Polleres, A.: Sparql 1.1 update. W3c recommendation, W3C (Mar 2013), \url{https://www.w3.org/TR/sparql11-update/}

\bibitem{groeger2021data}
Gr\"{o}ger, C.: There is no ai without data. Commun. ACM  \textbf{64}(11),  98–108 (Oct 2021). \doi{10.1145/3448247}

\bibitem{groth2013prov}
Groth, P., Moreau, L.: Prov-overview. an overview of the prov family of documents. W3c working group note, W3C (April 2013), \url{https://www.w3.org/TR/prov-overview/}

\bibitem{w3c-ssn-sosa}
Haller, A., Janowicz, K., Cox, S., et~al.: Semantic sensor network (ssn) ontology / sosa: Sensor, observation, sample, and actuator ontology. W3C Recommendation REC-vocab-ssn, W3C (Oct 2017), \url{https://www.w3.org/TR/vocab-ssn/}

\bibitem{herschel2017survey}
Herschel, M., Diestelk\"{a}mper, R., Ben~Lahmar, H.: A survey on provenance: What for? what form? what from? The VLDB Journal  \textbf{26}(6),  881–906 (Dec 2017). \doi{10.1007/s00778-017-0486-1}

\bibitem{herschel2016provenance}
Herschel, M., Hlawatsch, M.: Provenance: On and behind the screens. In: Proceedings of the 2016 International Conference on Management of Data. p. 2213–2217. SIGMOD '16, Association for Computing Machinery, New York, NY, USA (2016). \doi{10.1145/2882903.2912568}

\bibitem{jahnke2023catalog}
Jahnke, N.F., Otto, B.: Data catalogs in the enterprise: Applications and integration. Datenbank-Spektrum  \textbf{23}(2),  89--96 (2023). \doi{10.1007/s13222-023-00445-2}

\bibitem{kaefer2018asm4ld}
K\"{a}fer, T., Harth, A.: Rule-based programming of user agents for linked data. In: Proceedings of the Workshop on Linked Data on the Web (LDOW). CEUR-WS (2018), \url{http://ceur-ws.org/Vol-2073/\#article-05}

\bibitem{kaefer2018wild}
K\"{a}fer, T., Harth, A.: Specifying, monitoring, and executing workflows in linked data environments. In: The Semantic Web – ISWC 2018: 17th International Semantic Web Conference, Monterey, CA, USA, October 8–12, 2018, Proceedings, Part I. p. 424–440. Springer-Verlag, Berlin, Heidelberg (2018). \doi{10.1007/978-3-030-00671-6_25}

\bibitem{shacl-spec}
Knublauch, H., Kontokostas, D.: Shapes constraint language (shacl). W3c recommendation, W3C (Jul 2017), \url{https://www.w3.org/TR/shacl/}

\bibitem{kropshofer2025catalog}
Kropshofer, J., Schrott, J., Wöß, W., et~al.: A survey on the functionalities of data catalog tools. IEEE Access  \textbf{13},  83297--83319 (2025). \doi{10.1109/ACCESS.2025.3568542}

\bibitem{labadie2020fair}
Labadie, C., Legner, C., Eurich, M., et~al.: Fair enough? enhancing the usage of enterprise data with data catalogs. In: 2020 IEEE 22nd Conference on Business Informatics (CBI). vol.~1, pp. 201--210 (2020). \doi{10.1109/CBI49978.2020.00029}

\bibitem{w3c-prov-o}
Lebo, T., Sahoo, S., McGuinness, D., et~al.: {PROV-O: The PROV Ontology}. W3C Recommendation TR-prov-o, W3C (Apr 2013), \url{https://www.w3.org/TR/prov-o/}

\bibitem{markovic2019ep-plan}
Markovic, M., Garijo, D., Edwards, P.: Linking abstract plans of scientific experiments to their corresponding execution traces. In: Proceedings of the Third International Workshop on Capturing Scientific Knowledge (Sciknow 2019). CEUR-WS (Oct 2019)

\bibitem{owl-s}
Martin, D., Burstein, M., Hobbs, J., et~al.: {OWL‑S: Semantic Markup for Web Services}. {W3C Working Group Note}, {W3C} (Nov 2004), \url{https://www.w3.org/submissions/OWL-S/}

\bibitem{missier2013dprov}
Missier, P., Dey, S., Belhajjame, K., et~al.: {D-PROV}: Extending the {PROV} provenance model with {Workflow} structure. In: 5th USENIX Workshop on the Theory and Practice of Provenance (TaPP 13). USENIX Association, Lombard, IL (Apr 2013), \url{https://www.usenix.org/conference/tapp13/technical-sessions/presentation/missier}

\bibitem{dcat-ap}
{Publications Office of the European Union}: {DCAT Application profile for data portals in Europe (DCAT-AP)}, \url{https://op.europa.eu/en/web/eu-vocabularies/dcat-ap}, {Accessed: Nov 30 2025}

\bibitem{qudt}
QUDTOrganization: Fairsharing.org: Qudt; quantities, units, dimensions and types, \url{https://doi.org/10.25504/FAIRsharing.d3pqw7}, {Accessed: Nov 30 2025}

\bibitem{schilling2020cdo}
Schilling, R., Aier, S., Winter, R., et~al.: Design dimensions for enterprise-wide data management: A chief data officer’s journey. In: Proc. 53rd Hawaii International Conference on System Sciences (HICSS 53) (2020), \url{https://scholarspace.manoa.hawaii.edu/bitstream/10125/64456/0576.pdf}

\bibitem{vanhatalo2009process}
Vanhatalo, J., Völzer, H., Koehler, J.: The refined process structure tree. Data \& Knowledge Engineering  \textbf{68}(9),  793--818 (2009), \url{https://doi.org/10.1016/j.datak.2009.02.015}, sixth International Conference on Business Process Management (BPM 2008) – Five selected and extended papers

\bibitem{wilkinson2016fair}
Wilkinson, M.D., Dumontier, M., Aalbersberg, I.J., et~al.: The fair guiding principles for scientific data management and stewardship. Scientific Data  \textbf{3},  160018 (3 2016), \url{https://doi.org/10.1038/sdata.2016.18}

\end{thebibliography}

\appendix

\section{Further Figures}

\begin{figure}[h]
    \centering
    \includegraphics[width=0.75\linewidth]{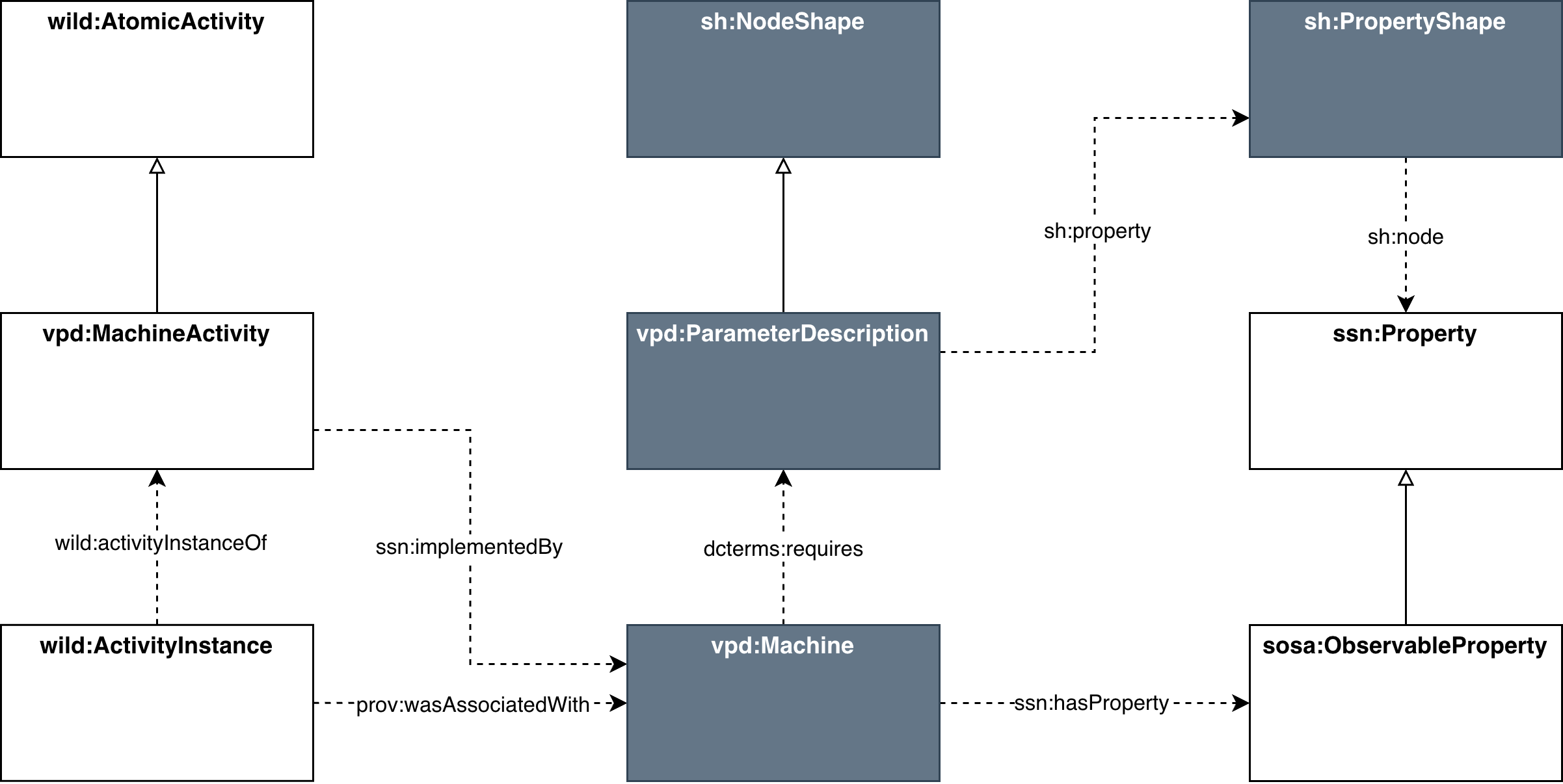}
    \caption{VPD-O view on parameter shapes.}
    \label{fig:vpd-o-parameter}
\end{figure}

\begin{figure}[h]
\centering
\includegraphics[width=0.85\textwidth]{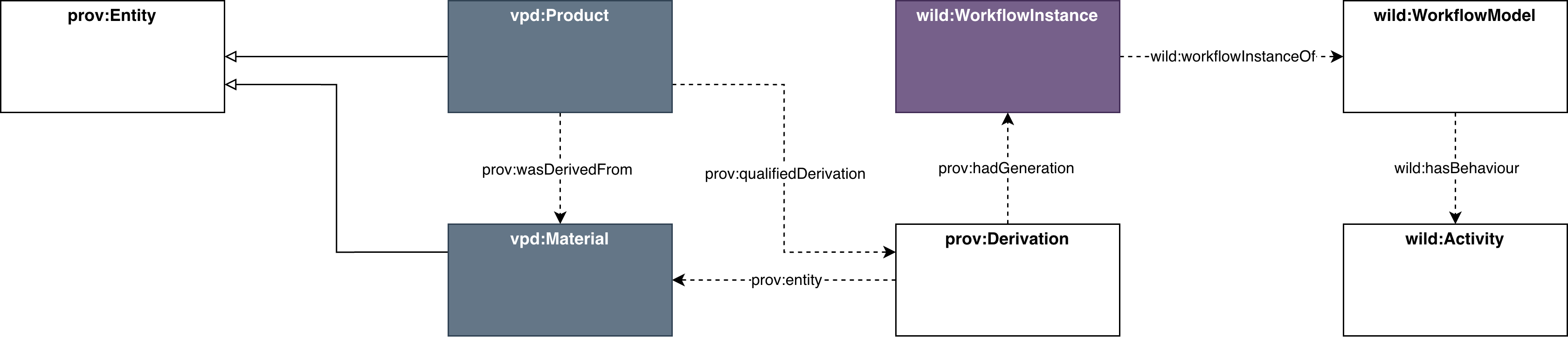}
\caption{VPD-O view on physical products.}
\label{fig:vpd-o-product}
\end{figure}

\end{document}